\documentstyle[eepic,aclap]{article}
\def\r{\mbox{~$\Rightarrow^*$}~~}
\def\W{W}
\def\output{\mbox{${\cal F}_{G,w}$}}
\def\GCNF{\mbox{$G$}}
\def\y{\mbox{``yes''}}
\def\n{\mbox{``no''}}
\def\myeps{\mbox{$\epsilon$}}

\newcommand{\ab}[1]{ \left \vert #1 \right \vert}
\newcommand{\set}[1]{ \{#1\}}
\newcommand{\qset}[2]{ \{{#1} : {#2}\}}
\newcommand{\tg}[1]{T(#1)}
\newcommand{\tw}[1]{t(#1)}

\newcommand{\cpa}[4]{ 
\begin{equation}
\renewcommand{\theequation}{#1}~~~~~~~~
{#2} \longrightarrow {#3},~~ {#4}
\end{equation}
}

\newcommand{\proof}[1]{
{\noindent {\it Proof.} {#1} \rule{2mm}{2mm} \vskip \belowdisplayskip}
}

\newtheorem{definition}{Definition}

\newtheorem{claim}{Claim}
\newtheorem{corollary}{Corollary}
\newtheorem{theorem}{Theorem}
\author{Lillian Lee \\ Division of Engineering and Applied Sciences \\
Harvard University \\ 33 Oxford Street \\ Cambridge, MA 012138 \\
{\tt llee@eecs.harvard.edu}
}

\title{\vspace{-75pt}
{\normalsize \tt \hfill Appears in the proceedings of ACL-EACL '97} \\ \mbox{}\\Fast Context-Free Parsing Requires Fast Boolean Matrix
Multiplication} 

\begin{document}
\bibliographystyle{fullname}

\maketitle
\begin{abstract}
Valiant showed that Boolean matrix multiplication (BMM) can be used
for CFG parsing.  We prove a dual result: CFG parsers running in time
$O(|G||w|^{3 - \myeps})$ on a grammar $G$ and a string $w$ can be used
to multiply $m \times m$ Boolean matrices in time $O(m^{3 -
\myeps/3})$.  In the process we also provide a formal definition of
parsing motivated by an informal notion due to Lang.  Our result
establishes one of the first limitations on general CFG parsing: a
fast, practical CFG parser would yield a fast, practical BMM
algorithm, which is not believed to exist.
\end{abstract}

\section{Introduction}

The context-free grammar (CFG) formalism was developed during the
birth of the field of computational linguistics.  The standard methods
for CFG parsing are the CKY algorithm \cite{Kasami:cky,Younger:cky}
and Earley's algorithm \cite{Earley:parse}, both of which have a
worst-case running time of $O(gN^3)$ for a CFG (in Chomsky normal
form) of size $g$ and a string of length $N$.  Graham et
al. \shortcite{Graham+Harrison+Ruzzo:journal} give a variant of
Earley's algorithm which runs in time $O(gN^3/\log N)$.  Valiant's
parsing method is the asymptotically fastest known \cite{Valiant:cfl}.
It uses Boolean matrix multiplication (BMM) to speed up the dynamic
programming in the CKY algorithm: its worst-case running time is $O(g
M(N))$, where $M(m)$ is the time it takes to multiply two $m \times m$
Boolean matrices together.

The standard method for multiplying matrices takes time $O(m^3)$.
There exist matrix multiplication 
 algorithms with time complexity $O(m^{3- \delta})$;  for instance,
Strassen's  has a worst-case running time of
$O(m^{2.81})$ \cite{Strassen:mult}, and the fastest currently known
has a worst-case running time
of $O(m^{2.376})$ \cite{Coppersmith+Winograd}. Unfortunately, the constants involved are so large that these
{\em fast} algorithms  (with the possible exception of Strassen's)
cannot be used in practice.
As matrix
multiplication is a very well-studied problem (see Strassen's
historical account \cite[section 10]{Strassen:hist}), it is highly
unlikely that simple, practical fast matrix multiplication algorithms
exist.  
Since the best BMM algorithms all rely on general matrix multiplication\footnote{   The ``four Russians''
algorithm \cite{four-russians}, the fastest BMM algorithm that does
not simply use ordinary matrix multiplication, has worst-case running
time $O(m^3/\log m)$. }, 
it is widely believed that there are no practical $O(m^{3 -  \delta})$
BMM algorithms.

One might therefore hope to find a way to speed up CFG parsing without
relying on matrix multiplication.  However, we show in this paper that
fast CFG parsing {\em requires} fast Boolean matrix multiplication in
a precise sense: any parser running in time $O(gN^{3 - \myeps})$ that
represents parse data in a retrieval-efficient way can be converted
with little computational overhead
into a $O(m^{3 - \myeps / 3})$ BMM algorithm.  Since it is very improbable
that practical fast matrix multiplication algorithms exist, we thus
establish one of the first nontrivial limitations on practical CFG
parsing. 

Our technique, adapted from that used by Satta \shortcite{Satta:bmm}
for tree-adjoining grammar (TAG) parsing, is to show that BMM can be
efficiently reduced to CFG parsing.  Satta's result does not apply to
CFG parsing, since it explicitly relies on the properties of TAGs that
allow them to generate non-context-free languages.
 
\section{Definitions}

A Boolean matrix is a matrix with entries from the set $\set{0,1}$.  A
Boolean matrix multiplication algorithm takes as input two $m
\times m$ Boolean matrices $A$ and $B$ and returns their {\em Boolean
product} $A \times B$, which is the $m \times m$ Boolean matrix $C$
whose entries $c_{ij}$ are defined by
$$c_{ij} = \bigvee_{k=1}^m \left( a_{ik} \wedge b_{kj} \right).$$
That is, $c_{ij} =1$ if and only if there exists a number $k$, $1 \leq k \leq m$,
such that $a_{ik} = b_{kj} = 1$.

We use the usual definition of a context-free grammar (CFG) as a
4-tuple $G=(\Sigma, V, R, S)$, where $\Sigma$ is the set of terminals,
$V$ is the set of nonterminals, $R$ is the set of productions, and $S \in V$
is the start symbol.  Given a string $w = w_1 w_2 \cdots w_N$ over
$\Sigma^*$, where each $w_i$ is an element of $\Sigma$, we use the
notation $w_i^j$ to denote the substring $w_i w_{i+1} \cdots w_{j-1} w_j$.

We will be concerned with the notion of {\em c-derivations}, which are
substring derivations that are consistent with a derivation of an
entire string.  Intuitively, $A \r w_i^j$ is a c-derivation
if it is consistent with at least one parse of $w$.

\begin{definition}
Let $G=(\Sigma, V, R, S)$ be a CFG, and let $w= w_1 w_2 \cdots w_N$,
$w_i \in \Sigma$.  A nonterminal
$A \in V$ {\em c-derives} (consistently derives) $w_i^j$ if and only
if the following conditions hold:
\begin{itemize}
\item $A \r w_i^j$, and
\item $S \r w_1^{i-1} A w_{j+1}^N.$
\end{itemize}
(These conditions together imply that $S \r w$.)  
\end{definition}

We would like our results to apply to all ``practical'' parsers, but
what does it mean for a parser to be practical?  First, we would like
to be able to retrieve constituent information for all possible parses
of a string (after all, the recovery of structural information is what
distinguishes parsing algorithms from recognition algorithms); such
information is very useful for applications like natural language
understanding, where multiple interpretations for a sentence may
result from different constituent structures.  Therefore, practical
parsers should keep track of c-derivations.  Secondly, a parser should
create an output structure from which information about constituents
can be retrieved in an efficient way --- \newcite{Satta:bmm} points
out an observation of Lang to the effect that one can consider the
input string itself to be a retrieval-inefficient representation of
parse information. In short, we require  practical parsers to output
a representation of the parse forest for a string that allows efficient
retrieval of parse information.  Lang in fact argues that parsing
means {\em exactly} the production of a shared forest structure ``from
which any specific parse can be extracted in time linear with the size
of the extracted parse tree'' \cite[pg. 487]{Lang:recog}, and
\newcite{Satta:bmm} makes this assumption as well.

These notions lead us to equate practical parsers with the class of
{\em c-parsers}, which keep track of c-derivations and may also
calculate general substring derivations as well.  
\begin{definition}
A c-parser is an algorithm that takes a CFG
grammar $G=(\Sigma, V, R, S)$ and string $w \in \Sigma^*$ as input and
produces output $\output$;  $\output$ acts as an oracle about parse
information, as follows:
\begin{itemize}
\item If $A$ c-derives $w_i^j$, then $\output(A,i,j) = \y$.
\item If $A \not \Rightarrow^* w_i^j$ (which implies that $A$ does not
c-derive $w_i^j$), then $\output(A,i,j) = \n.$
\item $\output$ answers queries in constant time.
\end{itemize}
\end{definition}
Note that the answer
$\output$ gives can be arbitrary if $A \r w_i^j$ but $A$ does not
c-derive $w_i^j$.  The constant-time constraint encodes the notion
that information extraction is efficient; observe that this is a
stronger condition than that called for by Lang.

We define c-parsers in this way to make the class of c-parsers as
broad as possible.  If we had changed the first condition to ``If $A$
derives $w_i^j \ldots$'', then Earley parsers would be excluded, since
they do not keep track of all substring derivations. If we
had written the second condition as ``If $A$ does not c-derive
$w_i^j$, then \ldots'', then CKY parsers would not be c-parsers, since
they keep track of all substring derivations, not just c-derivations.
So as it stands, the class of c-parsers includes tabular parsers
(e.g. CKY), where $\output$ is the table of substring derivations, and
Earley-type parsers, where $\output$ is the chart.  Indeed, it
includes all of the parsing algorithms mentioned in the introduction,
and can be thought of as a formalization of Lang's informal definition
of parsing.

\section{The reduction}

We will reduce BMM to c-parsing, thus proving that any c-parsing
algorithm can be used as a Boolean matrix multiplication algorithm.
Our method, adapted from that of \newcite{Satta:bmm} (who considered
the problem of parsing with tree-adjoining grammars), is to encode
information about Boolean matrices into a CFG.  Thus, given two
Boolean matrices, we need to produce a string and a grammar such that
parsing the string with respect to the grammar yields output
from which information about the product of the two matrices can be
easily retrieved.  

We can sketch the behavior of the grammar as follows.  Suppose entries
$a_{ik}$ in $A$ 
and $b_{kj}$ in $B$ are both 1.  Assume we have
some way to break up array indices into two parts so that $i$ can be
reconstructed from  $i_1$ and  $i_2$, 
$j$ can be
reconstructed from  $j_1$ and  $j_2$, and
$k$ can be
reconstructed from  $k_1$ and  $k_2$. (We will describe a way to do
this later.)
Then, we will have the following derivation (for a quantity $\delta$
to be defined later) :

\begin{eqnarray*}
C_{i_1,j_1} &\Rightarrow & A_{i_1,k_1} B_{k_1,j_1} \\
 &\Rightarrow^*& 
\underbrace{w_{i_2}  \cdots w_{k_2 + \delta}}_{\mbox{derived by $A_{i_1,k_1}$}} 
\underbrace{w_{k_2 + \delta + 1} \cdots w_{j_2 + 2
\delta}}_{\mbox{derived by $B_{k_1,j_1}$}}.
\end{eqnarray*}
The key thing to observe is that $C_{i_1,j_1}$ generates two
nonterminals whose ``inner'' indices match, and that these two
nonterminals generate substrings that lie exactly next to each other.
The ``inner'' indices constitute a check on $k_1$, and the substring
adjacency constitutes a check on $k_2$.

Let $A$ and $B$ be two Boolean matrices, each of size $m \times m$,
and let $C$ be their Boolean matrix product, $C=A \times B$.  In the
rest of this section, we consider $A$, $B$, $C$, and $m$ to be fixed.
Set $n = \lceil m^{1/3} \rceil$, and set $\delta = n +2$.  We will be
constructing a string of length $3 \delta$; we choose $\delta$
slightly larger than $n$ in order to avoid having epsilon-productions in our
grammar.

Recall that $c_{ij}$ is non-zero if and only if we can find a non-zero
$a_{ik}$ and a non-zero $b_{\overline{k} j}$ such that $k =
\overline{k}$.  In essence, we need simply check for the equality of
indices $k$ and $\overline{k}$.  We will break matrix indices into two
parts: our grammar will check whether the first parts of $k$ and $
\overline{k}$ are equal, and our string will check whether the second
parts are also equal, as we sketched above.  Encoding the indices 
ensures that the grammar is of as small a size as possible, which
will be important for our time bound results.

Our index encoding function is as follows.  Let $i$ be a matrix
index, $1 \leq i \leq m$.  Then we define the function $f(i) =
(f_1(i), f_2(i))$ by
\begin{eqnarray*}
f_1(i) & = & \lfloor i/n \rfloor ~~~~~~~~~~~~~  (0 \leq f_1(i) \leq
n^2), ~ \mbox{and} \\
f_2(i) & = & (i ~ {\rm mod} ~ n) + 2 ~~ (2 \leq f_2(i) \leq n+1).
\end{eqnarray*}
Since $f_1$ and $f_2$ are essentially the quotient and remainder of
integer division of $i$ by $n$, we can retrieve $i$ from $(f_1(i), f_2(i))$.
We will use the notational shorthand of using subscripts instead of
the functions $f_1$ and $f_2$, that is,  we write $i_1$ and $i_2$
for $f_1(i)$ and $f_2(i)$.

It is now our job to create a CFG $G = (\Sigma,V,R,S)$ and a string
$w$ that encode information about $A$ and $B$ and express
constraints about their product $C$.  Our plan is to
include a set of nonterminals $\qset{C_{p,q}}{1 \leq p, q \leq n^2}$
in $V$ so that $c_{ij} = 1$ if and only if $C_{i_1,j_1}$ c-derives
$w_{i_2}^{j_2 + 2 \delta}$.  In section \ref{grammar1}, we 
describe a version of $G$ and prove it has this c-derivation
property.  Then, in section \ref{CNF} we explain that $G$ can
easily be converted to Chomsky normal form in such a way as to
preserve c-derivations.

We choose the set of terminals to be 
$\Sigma = \qset{w_\ell}{1 \leq \ell \leq 3n + 6}$, and choose the
string to be parsed to be
$w = w_1 w_2 \cdots w_{3n+6}$.
We consider $w$ to be made up of three parts, $x$, $y$, and $z$, each of size $\delta$:  
$w = \underbrace{w_1 w_2 \cdots w_{n+2}}_{x} \underbrace{w_{n+3}
\cdots w_{2n+4}}_{y} \underbrace{w_{2n+5}   \cdots w_{3n+6}}_z$.
Observe that for any $i$, $1 \leq i \leq m$, $w_{i_2}$ lies within $x$, $w_{i_2 + \delta}$ lies within $y$,
and $w_{i_2 + 2 \delta}$ lies within $z$, since
\begin{eqnarray*}
 i_2 & \in & [2, n+1], \\
 i_2 + \delta & \in & [n+4, 2n+3], \mbox{ and} \\
i_2 + 2 \delta & \in & [2n + 6, 3n+5].
\end{eqnarray*}

\subsection{The grammar}
\label{grammar1}
Now we begin building the grammar $G=(\Sigma, V, R, S)$.  We start
with the nonterminals $V = \set{S}$
and the production set $R = \emptyset$.
We add nonterminal $\W$ to $V$ for generating arbitrary
non-empty substrings of $w$; thus we need the productions
\cpa{$\W$-\mbox{rules}}{\W}{w_\ell \W \vert w_\ell}{1 \leq \ell \leq 3n+6.}
Next we encode the entries of the input matrices $A$ and $B$ in our grammar.
We include sets of non-terminals $\qset{A_{p,q}}{1 \leq p, q \leq n^2}$
and $\qset{B_{p,q}}{1 \leq p, q \leq n^2}$.
Then, for every {\em non-zero} entry $a_{ij}$ in $A$, we add the
production
\begin{equation}
\renewcommand{\theequation}{$A$-\mbox{rules}} 
{A_{i_1,j_i}} \longrightarrow {w_{i_2} \W w_{j_2 + \delta}.}
\label{A-rules}
\end{equation}

For every {\em non-zero} entry $b_{ij}$ in $B$, we add the production
\begin{equation}
\renewcommand{\theequation}{$B$-\mbox{rules}} 
~~~~~~{B_{i_1,j_i}} \longrightarrow {w_{i_2 + 1 + \delta} \W w_{j_2 + 2 \delta}.}
\label{B-rules}
\end{equation}

We need to represent entries of $C$, so we
create nonterminals $\qset{C_{p,q}}{1 \leq p, q \leq n^2}$ and productions
\begin{equation}
\renewcommand{\theequation}{$C$-\mbox{rules}}
\hspace{.6in}{C_{p,q}} \longrightarrow {A_{p,r} B_{r,q}},~~ {1 \leq p,
q, r \leq n^2.}
\label{C-rules}
\end{equation}

Finally, we complete the construction with productions for the
start symbol $S$:
\cpa{$S$-\mbox{rules}}{S}{\W C_{p,q} \W}{1 \leq p, q \leq n^2.}

We now prove the following result about the grammar and string
we have just described.
\begin{theorem}
For $1 \leq i, j \leq m$, the entry 
$c_{ij}$ in $C$ is non-zero if and only if $C_{i_1,j_1}$
c-derives $w_{i_2}^{j_2 + 2 \delta}$.
\label{thm:correct}
\end{theorem}
\proof{
Fix $i$ and $j$.

\begin{figure*}
\begin{center}
\setlength{\unitlength}{0.00083333in}
\begingroup\makeatletter\ifx\SetFigFont\undefined
\def\x#1#2#3#4#5#6#7\relax{\def\x{#1#2#3#4#5#6}}%
\expandafter\x\fmtname xxxxxx\relax \def\y{splain}%
\ifx\x\y   
\gdef\SetFigFont#1#2#3{}%
\else
\gdef\SetFigFont#1#2#3{\begingroup
  \count@#1\relax \ifnum 25<\count@\count@25\fi
  \def\x{\endgroup\@setsize\SetFigFont{#2pt}}%
  \expandafter\x
    \csname \romannumeral\the\count@ pt\expandafter\endcsname
    \csname @\romannumeral\the\count@ pt\endcsname
  \csname #3\endcsname}%
\fi
\gdef\SetFigFont#1#2#3{}
\fi\endgroup
\begin{picture}(6673,3525)(0,-10)
\put(2945,3402){\makebox(0,0)[lb]{\smash{{{\SetFigFont{12}{14.4}{rm}$S$}}}}}
\path(2475,1827)(3000,2427)(3525,1827)
\put(2925,2502){\makebox(0,0)[lb]{\smash{{{\SetFigFont{12}{14.4}{rm}$C_{i_1,j_1}$}}}}}
\put(2325,1602){\makebox(0,0)[lb]{\smash{{{\SetFigFont{12}{14.4}{rm}$A_{i_1,k_1}$}}}}}
\put(3485,1602){\makebox(0,0)[lb]{\smash{{{\SetFigFont{12}{14.4}{rm}$B_{k_1,j_1}$}}}}}
\path(12,297)(2088,297)
\path(132.000,327.000)(12.000,297.000)(132.000,267.000)
\path(1968.000,267.000)(2088.000,297.000)(1968.000,327.000)
\path(2112,297)(4188,297)
\path(2232.000,327.000)(2112.000,297.000)(2232.000,267.000)
\path(4068.000,267.000)(4188.000,297.000)(4068.000,327.000)
\path(4212,297)(6288,297)
\path(4332.000,327.000)(4212.000,297.000)(4332.000,267.000)
\path(6168.000,267.000)(6288.000,297.000)(6168.000,327.000)
\put(900,27){\makebox(0,0)[lb]{\smash{{{\SetFigFont{12}{14.4}{rm}$x$}}}}}
\put(3000,27){\makebox(0,0)[lb]{\smash{{{\SetFigFont{12}{14.4}{rm}$y$}}}}}
\put(5325,27){\makebox(0,0)[lb]{\smash{{{\SetFigFont{12}{14.4}{rm}$z$}}}}}
\path(3000,3327)(3000,2652)
\path(900,2577)(2850,3327)
\texture{44555555 55aaaaaa aa555555 55aaaaaa aa555555 55aaaaaa aa555555 55aaaaaa 
	aa555555 55aaaaaa aa555555 55aaaaaa aa555555 55aaaaaa aa555555 55aaaaaa 
	aa555555 55aaaaaa aa555555 55aaaaaa aa555555 55aaaaaa aa555555 55aaaaaa 
	aa555555 55aaaaaa aa555555 55aaaaaa aa555555 55aaaaaa aa555555 55aaaaaa }

\shade\path(3000,702)(2400,1527)(1125,702)(3000,702)
\path(3000,702)(2400,1527)(1125,702)(3000,702)

\shade\path(3000,702)(3525,1527)(4800,702)(3000,702)
\path(3000,702)(3525,1527)(4800,702)(3000,702)
\path(5025,2652)(3150,3327)
\shade\path(3000,702)(2400,1527)(1125,702)(3000,702)
\path(3000,702)(2400,1527)(1125,702)(3000,702)
\texture{88555555 55000000 555555 55000000 555555 55000000 555555 55000000 
      555555 55000000 555555 55000000 555555 55000000 555555 55000000 
      555555 55000000 555555 55000000 555555 55000000 555555 55000000 
      555555 55000000 555555 55000000 555555 55000000 555555 55000000 }

\shade\path(1080,702)(30,702)(780,2382)
        (1110,687)(1080,702)
\path(1080,702)(30,702)(780,2382)
        (1110,687)(1080,702)

\shade\path(4815,702)(6390,702)(5295,2442)
      (4815,687)(4815,702)(4815,702)
\path(4815,702)(6390,702)(5295,2442)
      (4815,687)(4815,702)(4815,702)

\put(0,552){\makebox(0,0)[lb]{\smash{{{\SetFigFont{12}{14.4}{rm}$w_1$}}}}}
\put(3000,552){\makebox(0,0)[lb]{\smash{{{\SetFigFont{12}{14.4}{rm}$w_{k_2+1+\delta}$}}}}}
\put(2550,552){\makebox(0,0)[lb]{\smash{{{\SetFigFont{12}{14.4}{rm}$w_{k_2+\delta}$}}}}}
\put(4650,552){\makebox(0,0)[lb]{\smash{{{\SetFigFont{12}{14.4}{rm}$w_{j_2
+ 2\delta}$}}}}}
\put(1185,552){\makebox(0,0)[lb]{\smash{{{\SetFigFont{12}{14.4}{rm}$w_{i_2}$}}}}}
\put(675,2427){\makebox(0,0)[lb]{\smash{{{\SetFigFont{12}{14.4}{rm}$W$}}}}}
\put(1650,552){\makebox(0,0)[lb]{\smash{{{\SetFigFont{34}{40.8}{rm}...}}}}}
\put(375,552){\makebox(0,0)[lb]{\smash{{{\SetFigFont{34}{40.8}{rm}...}}}}}
\put(3975,552){\makebox(0,0)[lb]{\smash{{{\SetFigFont{34}{40.8}{rm}...}}}}}
\put(5400,552){\makebox(0,0)[lb]{\smash{{{\SetFigFont{34}{40.8}{rm}...}}}}}
\put(5175,2502){\makebox(0,0)[lb]{\smash{{{\SetFigFont{12}{14.4}{rm}$W$}}}}}
\put(6150,552){\makebox(0,0)[lb]{\smash{{{\SetFigFont{12}{14.4}{rm}$w_{3n+6}$}}}}}
\end{picture}

\end{center}
\caption{Schematic of the derivation process when $a_{ik} = b_{kj} =
1$.  The substrings derived by
$A_{i_1,k_1}$ and $B_{k_1,j_1}$ lie right next to each other.}
\label{schematic}
\end{figure*}
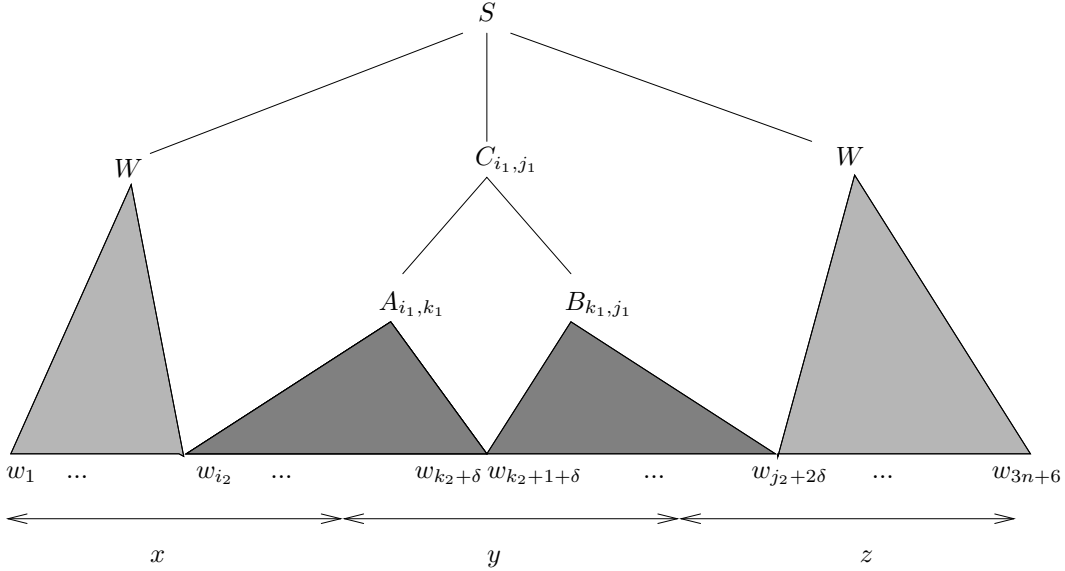

Let us prove the ``only if'' direction first.
Thus, suppose  $c_{ij} = 1$.  Then there exists a
$k$ such that $a_{ik} = b_{kj} = 1$.  Figure \ref{schematic} sketches how $C_{i_1,j_1}$ c-derives $w_{i_2}^{j_2 + 2 \delta}$.
\begin{claim}
$C_{i_1, j_1} \r w_{i_2}^{j_2 + 2 \delta}$.
\label{claim1}
\end{claim}
The production
$C_{i_1, j_1} \longrightarrow A_{i_1,k_1} B_{k_1,j_1}$ is
one of the $C$-rules in our grammar.
Since $a_{ik} = 1$,
$A_{i_1, k_1} \longrightarrow w_{i_2} \W w_{k_2 + \delta}$
is one of our $A$-rules, and since $b_{kj} = 1$,
$B_{k_1, j_1} \longrightarrow w_{k_2 + 1 + \delta} \W w_{j_2 + 2 \delta}$
is one of our $B$-rules.
Finally, since $i_2 + 1 < (k_2 + \delta) - 1 $ and 
$(k_2 + 1 + \delta) + 1 \leq (j_2 + 2 \delta) - 1$, we have
$\W \r w_{i_2 + 1}^{k_2 + \delta - 1}$ and
$\W \r w_{k_2 + 2 + \delta}^{j_2 + 2 \delta - 1}$,
since both substrings are of length at least one.
Therefore,
\begin{eqnarray*}
C_{i_1, j_1} &\Rightarrow{}{}{}{}{}{}{}&A _{i_1, k_1} B_{k_1, j_1} \\
 & \r &
\underbrace{w_{i_2} \W w_{k_2 + \delta}}_{\mbox{derived by } A_{i_1, k_1}}
\underbrace{ w_{k_2 + 1 + \delta} \W w_{j_2 + 2 \delta}}_{\mbox{derived
by } B_{k_1, j_1} } \\
 &\r& w_{i_2}^{j_2 + 2 \delta},
\end{eqnarray*}
and Claim~\ref{claim1} follows.

\begin{claim}
$S \r w_1^{i_2-1}C_{i_1,j_1}  w_{j_2 + 2 \delta + 1}^{3n +6}.$
\label{claim2}
\end{claim}
This claim is essentially trivial, since 
by the definition of the $S$-rules, we know that $S \r \W C_{i_1,
j_1} \W$. 
We need only show that neither $w_1^{i_2 - 1}$ nor $w_{j_2 + 2
\delta + 1}^{3n + 6}$ is the empty string (and hence can be derived by $\W$); since $1 \leq
i_2 - 1$ and $j_2 + 2 \delta + 1 \leq 3n + 6$, the claim holds.

Claims~\ref{claim1} and \ref{claim2} together
prove that $C_{i_1, j_1}$ c-derives $w_{i_2}^{j_2 + 2
\delta}$, as required.\footnote{This proof would have been simpler if we had
allowed $\W$ to derive the empty string.  However, we avoid epsilon-productions in order to facilitate the conversion to Chomsky normal
form, discussed later.}

Next we prove the ``if'' direction.
Suppose $C_{i_1, j_1}$ c-derives $w_{i_2}^{j_2 + 2 \delta}$,
which by definition means 
$C_{i_1, j_1} \r w_{i_2}^{j_2 + 2 \delta}$.
Then there must be a derivation resulting from the application of a $C$-rule as follows:
$$C_{i_1, j_1} \Rightarrow A_{i_1, k'} B_{k', j_1}
\r w_{i_2}^{j_2 + 2 \delta}$$
for some $k'$.
It must be the case that for some $\ell$,
$A_{i_1, k'} \r w_{i_2}^\ell$ and 
$B_{k', j_1} \r w_{\ell+1}^{j_2 + 2 \delta}$.
But then we must have
the productions
$A_{i_1, k'} \longrightarrow w_{i_2} \W w_{\ell}$
and $B_{k', j_1} \longrightarrow w_{\ell + 1} \W w_{j_2 + 2 \delta}$
with $\ell = k'' + \delta$ for some $k''$.
But we can only have such productions if there exists a number $k$
such that $k_1 = k'$, $k_2 = k''$,
$a_{ik} = 1$, and $b_{kj} = 1$; and this implies
that $c_{ij}=1$.
} 

Examination of the proof reveals that we have also shown the following
two corollaries.

\begin{corollary}
For $1 \leq i,j \leq m$, $c_{ij} = 1$ if and only if $C_{i_1,j_1} \r w_{i_2}^{j_2 + 2 \delta}$.
\label{derive}
\end{corollary}

\begin{corollary}
$S \r w$ if and only if $C$ is not the all-zeroes matrix.
\end{corollary}

Let us now calculate the size of $G$.  $V$ consists of
$O((n^2)^2)=O(m^{4/3})$ nonterminals.  $R$ contains $O(n)$ $W$-rules
and $O((n^2)^2) = O(m^{4/3})$ $S$-rules.  There are at most $m^2$
$A$-rules, since we have an $A$-rule for each non-zero entry
in $A$; similarly, there are at most $m^2$ $B$-rules.  And lastly,
there are $(n^2)^3 = O(m^2)$ $C$-rules.  Therefore, our grammar is of
size $O(m^2)$; since $G$ encodes matrices $A$ and $B$, it is of
optimal size. 

\subsection{Chomsky normal form}
\label{CNF}
We would like our results to be true for the largest class of parsers
possible.  Since some parsers require the input grammar to be in
Chomsky normal form (CNF), we therefore wish to construct a CNF
version $G'$  of $G$.  However, in order to preserve time bounds, we
desire that $O(\ab{G'}) = O(\ab{G})$, and we also require that Theorem
\ref{thm:correct} holds for $G'$ as well as $G$.  

The standard algorithm for converting CFGs to CNF can yield a
quadratic blow-up in the size of the grammar and thus is clearly
unsatisfactory for our purposes.  However, since $G$ contains no
epsilon-productions or unit productions, it is easy to see that we can
convert $G$ simply by introducing a small ($O(n)$) number of
nonterminals without changing any c-derivations for the $C_{p,q}$.
Thus, from now on we will simply assume that $G$ is in CNF.

\subsection{Time bounds}

We are now in a position to prove our relation between time bounds
for Boolean matrix multiplication and time bounds for CFG parsing.  

\begin{theorem} Any c-parser $P$ with running time $O(\tg{g} \tw{N})$
on grammars of size $g$ and strings of length $N$ can be converted
into a BMM algorithm $M_P$ that runs in time $O( \max ( m^2, \tg{m^2} 
\tw{m^{1/3}}))$.  In particular, if $P$ takes time $O(gN^{3
-\myeps})$, then $M_P$ runs in time $O(m^{3 - \myeps / 3})$.
\label{time}
\end{theorem}
\proof{$M_P$ acts as follows.  Given two Boolean $m \times m$ matrices
$A$ and $B$, it constructs $\GCNF$ and $w$ as described above.  It feeds
$\GCNF$  and $w$ to $P$, which outputs $\output$.  To compute the product
matrix $C$, $M_P$ queries for each $i$ and $j$, $ 1 \leq i,j \leq m$,
whether $C_{i_1, j_1}$ derives $w_{i_2}^{j_2 + 2 \delta}$ (we do not
need to ask whether  $C_{i_1, j_1}$ c-derives $w_{i_2}^{j_2 + 2
\delta}$ because of corollary \ref{derive}), setting
$c_{ij}$ appropriately.  By definition of c-parsers, each such query
takes constant time.
Let us compute the running time of $M_P$.  It takes $O(m^2)$ time to
read the input matrices.  Since $\GCNF$ is of size $O(m^2)$ and
$\ab{w} = O(m^{1/3})$, it takes $O(m^2)$ time to build the input to
$P$, which then computes $\output$ in time $O(\tg{m^2} \tw{m^{1/3}})$.
Retrieving $C$ takes $O(m^2)$.  So the total time spent by $M_P$ is
$O(\max (m^2, \tg{m^2} \tw{m^{1/3}}))$, as was claimed.

In the case where $\tg{g} = g$ and $\tw{N} = N^{3 -\myeps}$, $M_P$
has a running time of $O(m^2 (m^{1/3})^{3 - \myeps}) = O(m^{2 + 1 -
\myeps /3}) = O(m^{3 -\myeps /3})$.
} 

The case in which $P$ takes time linear in the grammar size is of the
most interest, since in natural language processing applications, the
grammar tends to be far larger than the strings to be parsed.  Observe
that theorem \ref{time} translates the running time of the standard CFG
parsers, $O(gN^3)$, into the running time of the standard BMM
algorithm, $O(m^3)$.  Also, a c-parser with running time $O(gN^{2.43})$
would yield a matrix multiplication algorithm rivalling that of
Strassen's, and a c-parser with running time better than $O(gN^{1.12})$
could be converted into a BMM method faster than Coppersmith and
Winograd.  As per the discussion above, even if such parsers exist,
they would in all likelihood not be very practical.  Finally, we note
that if a lower bound on BMM of the form $\Omega(m^{3 - \alpha})$ were found,
then we would have an immediate lower bound of $\Omega( N^{3 - 3
\alpha})$ on c-parsers running in time linear in $g$.


\section{Related results and conclusion}

We have shown that fast practical CFG parsing algorithms yield fast
practical BMM algorithms. Given that fast practical BMM algorithms are
unlikely to exist, we have established a limitation on practical CFG
parsing.  

Valiant (personal communication) notes that there is a reduction of $m
\times m$ Boolean matrix multiplication checking to context-free
recognition of strings of length $m^2$; this reduction is alluded to
in a footnote of a paper by \newcite{Harrison+Havel}.  However, this
reduction converts a parser running in time $O(|w|^{1.5})$ to a BMM
checking algorithm running in time $O(m^3)$ (the running time of the
standard multiplication method), whereas our result says that
sub-cubic practical parsers are quite unlikely; thus, our result is
quite a bit stronger.

\newcite{Seiferas:lower} gives a simple proof of an
$\Omega(\frac{N^2}{\log N})$ lower bound (originally due to
\newcite{Gallaire}) for the problem of on-line linear CFL recognition
by multitape Turing machines.  However, his results concern on-line
recognition, which is a harder problem than parsing, and so do not
apply to the general off-line parsing case.

Finally, we recall Valiant's reduction of CFG parsing to boolean
matrix multiplication \cite{Valiant:cfl}; it is rather pleasing to have
the reduction cycle completed.

\section{Acknowledgments}

I thank Joshua Goodman, Rebecca Hwa, Jon Kleinberg, and Stuart Shieber
for many helpful comments and conversations. Thanks to Les Valiant for
pointing out the ``folklore'' reduction.  This material is based upon
work supported in part by the National Science Foundation under Grant
No. IRI-9350192.  I also gratefully acknowledge partial support from
an NSF Graduate Fellowship and an AT\&T GRPW/ALFP grant.
Finally, thanks to Giorgio Satta, who mailed me a preprint of his
BMM/TAG paper several years ago.

\end{document}